\documentclass[sigconf]{acmart}
\pdfoutput=1
\preto{\abstractkeywords}{\nolinenumbers} 
\AtBeginDocument{%
  \providecommand\BibTeX{{%
    \normalfont B\kern-0.5em{\scshape i\kern-0.25em b}\kern-0.8em\TeX}}}

\acmConference[Conference acronym 'XX]{Make sure to enter the correct
  conference title from your rights confirmation emai}{June 03--05,
  2018}{Woodstock, NY}

\acmPrice{15.00}
\acmISBN{978-1-4503-XXXX-X/18/06}
\usepackage{graphicx}
\usepackage{svg}
\usepackage{url}
\usepackage{float}
\usepackage{booktabs}
\usepackage{multirow}
\usepackage{arydshln}
\usepackage{enumitem}
\newcommand{\mydash}{\textendash\textendash}
\begin{document}

\title{Aligning LLMs through Multi-perspective User Preference Ranking-based Feedback for Programming Question Answering }


\author{Hongyu Yang}
\affiliation{%
  \institution{School of Computer Science and Technology, University of Science and Technology of China}
  \city{Hefei}
  \country{China}
 }
 \email{hongyuyang@mail.ustc.edu.cn}
 
\author{Liyang He}
\affiliation{%
  \institution{School of Computer Science and Technology, University of Science and Technology of China}
  \city{Hefei}
  \country{China}
 }
 \email{heliyang@mail.ustc.edu.cn}
 
\author{Min Hou}
\affiliation{%
  \institution{
Hefei University of Technology} 
 \city{Hefei}
  \country{China}
}
\email{minho@mail.ustc.edu.cn}

\author{Shuanghong Shen}
\affiliation{%
  \institution{School of Artificial Intelligence and Data Science, University of Science and Technology of China}
  \city{Hefei}
  \country{China}
 }
 \email{closer@mail.ustc.edu.cn}
 
\author{Rui Li}
\affiliation{%
  \institution{School of Computer Science and Technology, University of Science and Technology of China}
  \city{Hefei}
  \country{China}
 }
 \email{ruili2000@mail.ustc.edu.cn}
 
 \author{Jiahui Hou*}
\affiliation{%
  \institution{School of Computer Science and Technology, University of Science and Technology of China}
  \city{Hefei}
  \country{China}
 }
 \email{jhhou@ustc.edu.cn}
 
  \author{Jianhui Ma}
\affiliation{%
  \institution{School of Computer Science and Technology, University of Science and Technology of China}
  \city{Hefei}
  \country{China}
 }
 \email{jianhui@ustc.edu.cn}
 
\author{Junda Zhao}
\affiliation{%
  \institution{University of Toronto}
  \city{Toronto}
  \country{Canada}
 }
 \email{junda.zhao@mail.utoronto.ca}







\begin{abstract}
Code Community Question Answering (CCQA) seeks to tackle programming-related issues, thereby boosting productivity in both software engineering and academic research. Recent advancements in Reinforcement Learning from Human Feedback (RLHF) have transformed the fine-tuning process of Large Language Models (LLMs) to produce responses that closely mimic human behavior. Leveraging LLMs with RLHF for practical CCQA applications has thus emerged as a promising area of study.
Unlike standard code question-answering tasks, CCQA involves multiple possible answers, with varying user preferences for each response. Additionally, code communities often show a preference for new APIs. These challenges prevent LLMs from generating responses that cater to the diverse preferences of users in CCQA tasks.
To address these issues, we propose a novel framework called \textbf{A}ligning \textbf{L}LMs through \textbf{Mu}lti-perspective User \textbf{P}reference Ranking-based Feedback for Programming \textbf{Q}uestion \textbf{A}nswering (ALMupQA) to create user-focused responses. Our approach starts with Multi-perspective Preference Ranking Alignment (MPRA), which synthesizes varied user preferences based on the characteristics of answers from code communities. We then introduce a Retrieval-augmented In-context Learning (RIL) module to mitigate the problem of outdated answers by retrieving responses to similar questions from a question bank. Due to the limited availability of high-quality, multi-answer CCQA datasets, we also developed a dataset named StaCCQA from real code communities. Extensive experiments demonstrated the effectiveness of the ALMupQA framework in terms of accuracy and user preference. Compared to the base model, ALMupQA showed nearly an 11\% improvement in BLEU, with increases of 20\% and 17.5\% in BERTScore and CodeBERTScore, respectively.
\end{abstract}




\maketitle
\section{Introduction}
\begin{figure*}[ht]
    \includegraphics[width=0.96 \textwidth]{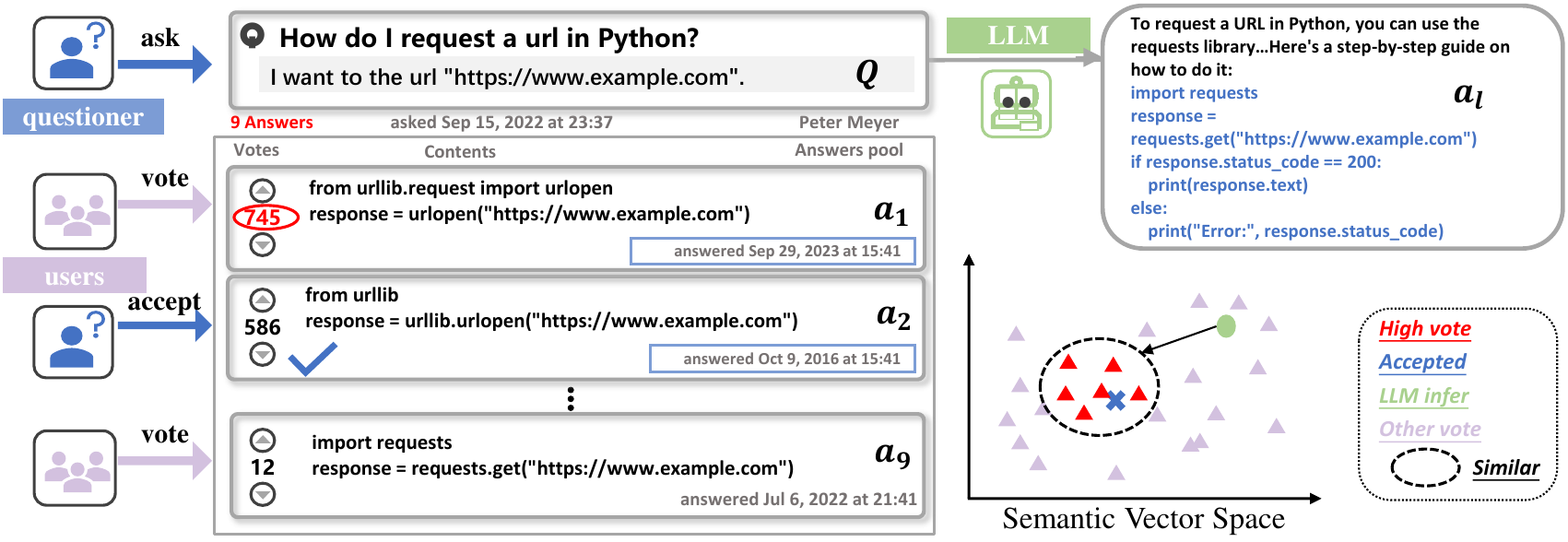}
    \caption{An example of a Code Community Question Answering. It encompasses key elements: a question \(Q\), a pool of answers \(\{a_1,\cdots,a_9\}\). Each \(a_i\) contains its text of content, the number of votes, and a label indicating whether the answer has been accepted by the questioner. Additionally, in the semantic vector space, there exists a certain distance between the LLM-based answers \(a_l\), the questioner-accepted answer \(a_2\), and the users-preferred answers \(a_1\).}
    \label{intro}
\end{figure*}
Large Language Models (LLMs) have demonstrated their success in the field of open-domain Question Answering (QA) \cite{nair2023drilling,openai2023gpt4,anil2023palm}. 
To align LLMs to users preference in domain-specific QA tasks, Reinforcement Learning from Human Feedback (RLHF) enables the alignment for human-like response generation \cite{ouyang2022training}. For example, in code question answers \cite{rafailov2023direct}, the LLM may produce redundant responses (e.g.,\(a_l\) in Figure~\ref{intro}). Utilizing RLHF can effectively achieve precise generation behavior control. However, the application of LLMs to real-world Code Community Question Answering (CCQA) tasks and the related preference alignment research remains an underexplored domain.

In recent years, CCQA has gained increasing significance in both academia \cite{nasehi2012makes,liu2021codeqa,zhou2018recurrent,chen2017enhancing} and industry \cite{diamantopoulos2015employing, amancio2021recency,ragkhitwetsagul2019toxic,kasela2023se}. It focuses on the code question-answer interactions among users in code communities (e.g., Stack Overflow\footnote{https://stackoverflow.com}). Unlike conventional QA task \cite{srivastava2019adapting,lin2021multi, yin2015answering}, CCQA exhibits three distinct characteristics. First, a question typically does not have just one answer, and as indicated in Table~\ref{data}, nearly 46\% of questions receive more than two answers, with some boasting an answer pool as large as 30. Second, each answer encompasses not only the textual content but also additional interactive elements, such as votes from other users, which reflect rich user preferences. Third, different users exhibit varying preferences for different answers to a given question. For example, in Figure \ref{intro}, a questioner posed a question \( Q \) and accepted answer \( a_2 \) from the pool of answers \( \{a_1, \cdots, a_9\} \), while some users favored answer \( a_1 \) with the highest votes.

Based on the characteristics mentioned above, to enable LLMs to be effectively applied in CCQA tasks and generate responses that satisfy diverse user preferences, several critical problems still need to be addressed. First, some works \cite{roziere2023code} have attempted to align LLM responses with human preferences by using the accepted answer (e.g., $a_2$ in Figure~\ref{intro}) as the alignment target. However, the accepted answer may not reflect the preferences of all users, as the answer chosen by the questioner may not be favored by other users. Second, although some studies \cite{zhou2018recurrent,chen2017enhancing,maia2021tag,du2021towards} have begun to focus on entire answers and have introduced content-based ranking methods, none have yet considered the inherent preferences of diverse users in CCQA and the feedback from LLMs. Third, it is worth noting that people's preferences shift with API updates in code communities, as they tend to choose newer versions of APIs. However, the accepted answer may suffer from being outdated, as in the field of programming, API updates occur rapidly. For example, in Figure~\ref{intro}, the "urllib" API in answer $a_2$ is applicable to Python 2 but deprecated in Python 3.

To overcome the above problems and limitations, we propose a novel multi-perspective preference ranking method for aligning LLMs on CCQA, which we call ALMupQA. ALMupQA primarily comprises two modules: Multi-perspective Preference Ranking Alignment (MPRA) and Retrieval-augmented In-context Learning (RIL). In MPRA, we first propose three scores as preference ratings for the answers, including a questioner-perspective bias score to assess the discrepancy between the accepted answer and other answers, a users-perspective vote score to reflect the collective preferences of other users, and a LLMs-perspective content score for evaluating the semantic quality of the answer content. Then, we introduce a preference ranking alignment method to utilize the three scores iteratively to identify the preference order of answers and optimize the alignment with user preferences using a list-wise contrastive loss. Besides, RIL aims to address the issue of outdated answers by retrieving answers to similar questions from the question bank and employing them as few-shot examples to enhance the effectiveness of the generated responses. Finally, due to the current lack of relevant datasets, we constructed a high-quality dataset, StaCCQA \footnote{Our dataset is accessible on https://anonymous.4open.science/r/PETCoQA-810A.}, from real-world code communities. Extensive experiments validated the effectiveness of the ALMupQA method in terms of accuracy and user preference. In summary, the paper makes three main contributions:
\begin{itemize}[leftmargin=*]
   \item  We propose a novel method, ALMupQA, to achieve preference alignment in the multi-perspective Community Code Question Answering task, which is an industrial practice with practical applications.
  \item In ALMupQA, we introduce the MPRA method to align preferences from a ranking perspective, taking into account the unique characteristics of CCQA answers, and propose RIL to address the issue of potentially outdated code.
  \item We constructed a multi-user preference dataset, StaCCQA, from the real-world code community. Comprehensive experiments on this dataset have evaluated the performance of ALMupQA against other open-source and proprietary LLM baselines. The results demonstrate the superiority of ALMupQA, establishing it as a robust foundational model for CCQA research.
\end{itemize}
\begin{figure*}[h]
    \includegraphics[width=0.92 \textwidth]{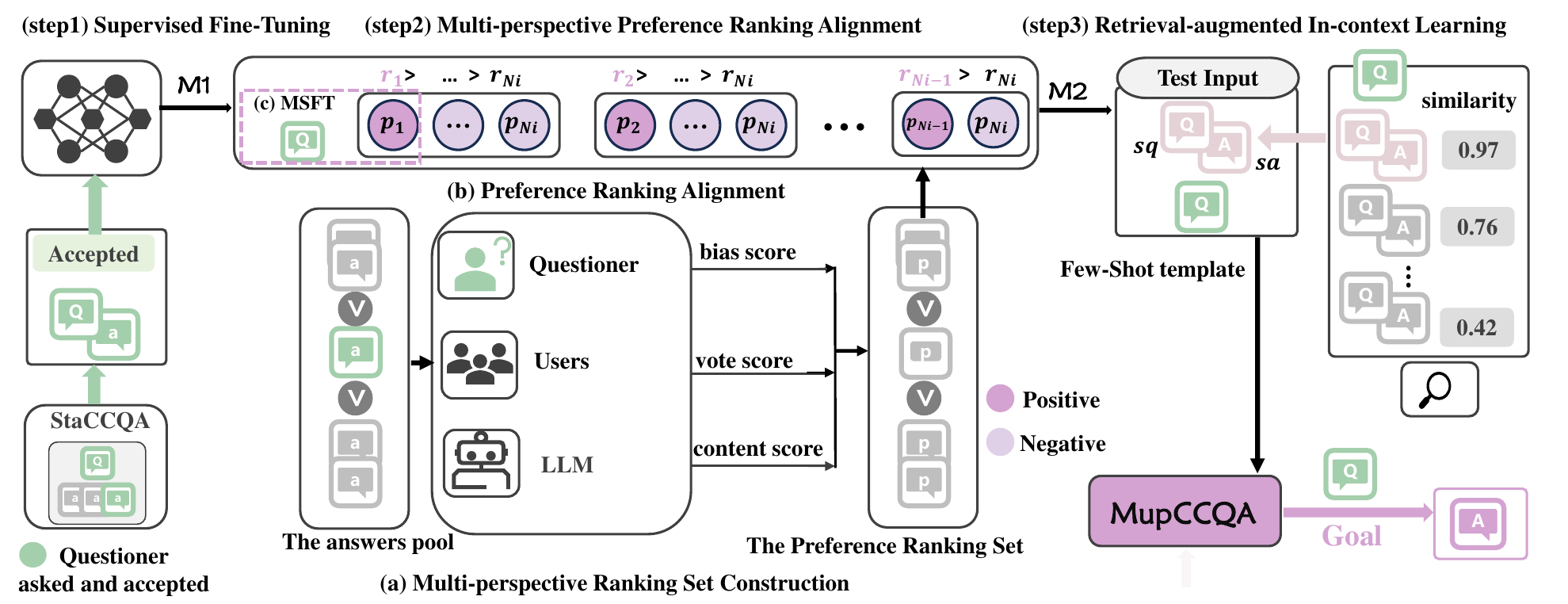}
    \caption{Overall architecture of the ALMupQA framework, including three stages: (step1) foundational Supervised Fine-Tuning (SFT) for acquiring programming-specific knowledge, (step2) Multi-perspective Preference Ranking Alignment (MPRA) for integrating diverse preferences, and (step3) Retrieval-augmented In-context Learning (RIL) to address the issue of outdated answers by retrieving the most similar post-solution pairs as prompts.}
    \label{model}
\end{figure*}
\section{Related works}
\subsection{Code Community Question Answering}
In code communities, programmers can both seek and share expertise, exemplifying the trend of collaborative problem-solving and knowledge exchange in software development. Code Community Question Answering (CCQA) is a fundamental task in code communities, which involves programming issues generated from user-posted questions and relevant answers to these questions \cite{liu2011predicting}.
Given the substantial differences between structured code and text, CCQA systems must possess the ability to comprehend both programming and natural languages, rendering this task highly challenging \cite{liu2021codeqa}. 

Within CCQA, we can identify numerous research topics, such as predicting answerable questions \cite{asaduzzaman2013answering}, assessing answer quality \cite{ragkhitwetsagul2019toxic, zhang2019empirical, gao2020generating}, answer generation \cite{zhou2018recurrent}, and answer ranking \cite{amancio2021recency, ginsca2013user, liu2021codeqa,dalip2013exploiting}. 
These answer ranking methods typically employ classical deep-learning models to utilize the answer text \cite{zhou2018recurrent} and the fundamental characteristics of the user \cite{ginsca2013user}.
For instance, L2R \cite{dalip2013exploiting} followed a learning to rank approach based on different groups of features like features referred to the users, stylistic or structural features. 
RCNN \cite{zhou2018recurrent} employed Gated Recurrent Units (GRU) with thread-level features for ranking answers.
Other research \cite{amancio2021recency} utilized recency and quality as criteria for ranking responses. However, few study has considered the inherent preferences of diverse users and LLM feedback. Therefore, exploring the ranking of answers based on preferences through the utilization of LLMs for alignment is a worthy endeavor.
\subsection{Preference Alignment for Question Answering}
In recent years, large language models (LLMs) \cite{openai2023gpt4,anil2023palm,zeng2022glm,llama1,llama2} have driven increasingly diverse applications, demonstrating notable expertise in question answering. By fine-tuning on extensive datasets across various programming domains, LLMs have also attained proficiency in synthesizing programs that are both syntactically correct and functionally accurate \cite{chen2021evaluating,nijkamp2022codegen,zheng2023codegeex,li2023starcoder,wang2023codet5+,roziere2023code}. This capability enables them to adeptly navigate the complexities of programming problems, including conceptual understanding, code generation, API utilization, and debugging \cite{hoq2023sann,phan2023evaluating,du2021single}.

Recently, reinforcement learning from human feedback (RLHF) \cite{stiennon2020learning,ouyang2022training,roziere2023code} has emerged as a milestone method for aligning with human preferences. This approach typically employs the Bradley-Terry model to optimize the neural network's reward function, followed by fine-tuning the language model using reinforcement learning algorithms, most commonly proximal policy optimization (PPO) \cite{schulman2017proximal}, to maximize the given reward.
Moreover, due to the sensitivity of RL parameters and the complex three-stage process of RLHF, numerous preference alignment methods have been proposed. For instance, RRHF \cite{yuan2023rrhf} introduced a boundary ranking loss function to optimize LLMs without requiring an additional reward model. DPO \cite{rafailov2023direct} introduced a direct preference optimization method, treating LLMs themselves as the reward model. PRO \cite{song2023preference} optimizes complex preference data through a listwise ranking loss function.
Crucially, LLMs exhibit their unique stylistic preferences in content generation, adeptly leveraging retrieved knowledge from prompts. Inspired by these insights, we propose aligning with human preferences through multi-perspective preference scoring by iteratively ranking the preference scores of all answers to a given question, rather than aligning preferences via a reward model.
\section{PRELIMINARIES}
\subsection{Task Formulation}
Our overall target is to design a multi-perspective preference alignment to guide a Large Language Model (LLM), denoted as $\mathcal{M}$, in generating answers that synthesize diverse user preferences with a real-world code community question answering dataset \(\mathcal{D} = \left\{ (q_i, \{a_1^i, a_2^i, \ldots, a_{N_i}^i\}) \mid i \in \{1, 2, \ldots, N\} \right\}\).
Here, $q_i$ represents the $i^{\text{th}}$ question, and \( \{a_1^i, a_2^i, \ldots, a_{N_i}^i\}\) represents the pool of answers for $q_i$. We denote $a = (c, v, a_c)$, with $c$ being the content of answer $a$; $v$ being the votes for answer $a$; and $a_c \in\{0,1\}$ representing whether the answer $a$ is accepted by the questioner. 
Formally, any \( q \) or \( c \) is a sequence of tokens, denoted as \( t = \{ t_i \mid t_i \in \mathcal{C}  \text{ or } t_i \in \mathcal{T} \} \), where \( t_i \) denotes the \( i \)-th token in the set \( t \), \( \mathcal{C} \) represents the set of code, and \( \mathcal{T} \) represents the set of text.

\subsection{Reinforcement Learning from Human Feedback}
We begin with a brief introduction to Reinforcement Learning from Human Feedback (RLHF)~\cite{ouyang2022training}, which primarily comprises three stages. The first stage is supervised fine-tuning on a LLM, denoted as $\mathcal{M}$, which is also a component of our framework and will be elaborated in Section~\ref{sec:sfp}. The second stage involves using the SFT model $\mathcal{M}_1$ to generate pairs of responses for a given prompt $\mathcal{I}$. These pairs have a preference order,  as illustrated by \(p_i\) is preferred over \(p_j\) in Figure \ref{method} (b).
 To predict these pairs, current works typically employ the Bradley-Terry (BT) model, which defines the preference probability as follows:
 \begin{equation}
    \mathcal{P}_{BT}=\frac{exp(r_{\phi}(\mathcal{I}_1,p_i) )}{exp(r_{\phi}(\mathcal{I}_1,p_i)) +exp(r_{\phi}(\mathcal{I}_1,p_j)) } 
    \label{bt}
\end{equation}
Where \(r_{\phi}\) is inherently a binary classification reward model, and \(\mathcal{I}_1\) is a QA prompt containing the question \(q\). The optimization objective of this stage is defined as a binary classification problem to train the reward model:
\begin{equation*}
    \mathcal{L}_{BT}=-log \sigma (r_{\Phi }(\mathcal{I}_1,p_i)-r_{\Phi }(\mathcal{I}_1,p_j))
\end{equation*}
In the third stage, RLHF leverages the acquired \(r_{\phi}\) to provide feedback to \(\mathcal{M}_1\) and \(\sigma \) is the logistic function. Specifically, the optimization problem of RLHF is formulated the following :
\begin{equation*}
    \max_{\mathcal{M}_2}\mathbb{E} (r_{\Phi}(\mathcal{I}_1,p)-\xi log\frac{\mathcal{M}_2(p|\mathcal{I}_1)}{\mathcal{M}_1(p|\mathcal{I}_1)}  )
\end{equation*}
In this context, the role of \(\xi \) is to regulate the deviation from the baseline reference policy \(\mathcal{M}_1\), ensuring diversity in the generated outputs and preventing the production of high-reward yet nonsensical answers. It is worth noting that RLHF generates pairs of responses, which is not enough to questions with more than two answers. Therefore, we need to explore a new method to adapt.

\section{Methodology}
As shown in Figure~\ref{model}, the ALMupQA framework encompasses three stages: (1) foundational Supervised Fine-Tuning to adapt to CCQA, (2) Multi-perspective Preference Ranking Alignment (MPRA) to integrate diverse preferences, and (3) Retrieval-augmented In-context Learning to address the issue of outdated answers.
\subsection{Foundational Supervised Fine-Tuning}
\label{sec:sfp}
Foundational LLMs are typically trained on open-domain corpora. To adapt these universal LLMs to programming-specific code community corpora, we first employ a Supervised Fine-Tuning \cite{ouyang2022training}. Specifically, we denote \(a^i\) with \(a_c = 1\) as \(a_c^i\) and select pairs \( (q_i, a_c^i) \) from the dataset \( \mathcal{D} \) with votes \( v^i \) exceeding 100 to form the training and validation dataset. Then we optimize the LLM as follows:
\begin{equation}
    \mathcal{ L}_ {SFT} =-\frac{1}{|a_c^i|} \sum_{j=1}^{|a_c^i|} log \mathcal{P_M}(a_c^{(i,j)}|\mathcal{I},q_i,a_c^{(i,<j) })
    \label{qpsft}
\end{equation}
where \(a_c^{(i,j)}\) is the \(j\)-th token of \(a_c^i\), \(\mathcal{I}\) is the prompt template, and \(\mathcal{P_M}\) denotes the token probability predicted by the model \(\mathcal{M}\). With this training objective, the QA data serves as the fundamental supervision information to fine-tune the model \(\mathcal{M}\) for the programming-specific QA scenario, resulting in a model denoted as \(\mathcal{M}_1\).
\begin{figure}[t]
    \vspace{-0.3cm}
    \includegraphics[width=\linewidth]{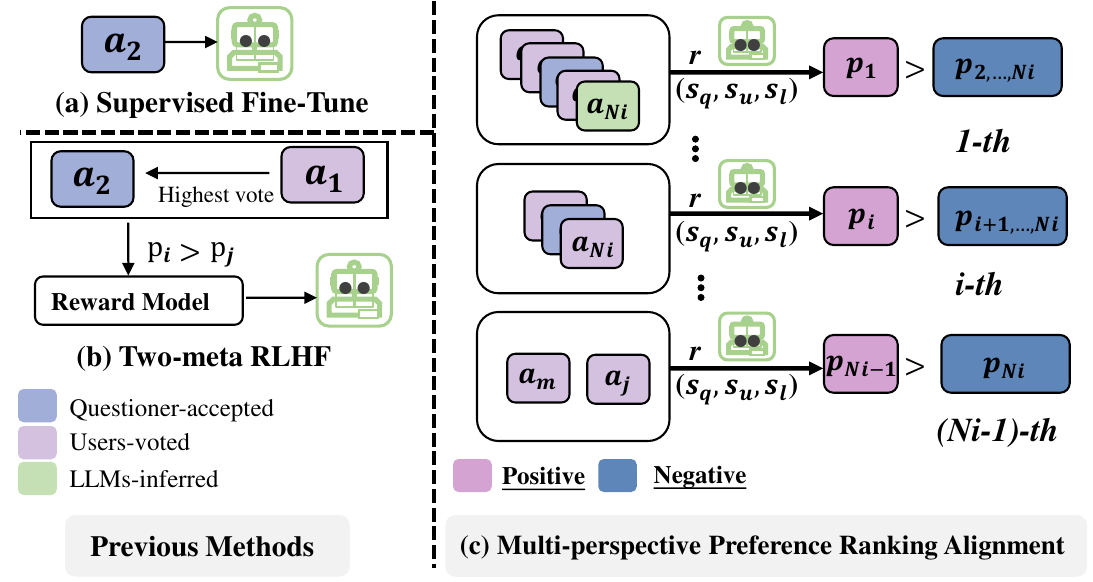}
    \caption{In CCQA, we compared previous human alignment methods with our approach. (a) SFT aligns only the answer accepted by the questioner $a_2$, while (b) RLHF compares $a_2$ with the highest-voted users-preferred answer $a_1$, sampling two-meta candidates $p_i \succ p_j$ from the entire ranking to train a reward model, and then relies on this reward model to fine-tune the base LLM. (c) Ours contrasts $p_i$ with all members in the preference set \(\{ p_1,\cdots,p_{N_i} \}\), based on the overall preference score $r$, which includes bias scores \(s_q\), vote scores \(s_u\), and content scores \(s_l\).  }
    \label{method}
    \vspace{-0.5cm}
\end{figure}
\subsection{Multi-perspective Preference Ranking Alignment}
To align multi-perspective preferences from different users, we introduce the Multi-perspective Preference Ranking Alignment (MPRA) method. First, we propose three distinct metric scores from different aspects to build the ranking set. Then, these composite scores act as the reward value within the preference ranking alignment in a list-wise contrastive learning fashion.
\subsubsection{\textbf{Multi-perspective Ranking Set Construction}}
First, as the answer chosen by the questioner may not be favored by other users\cite{kasela2023se}, we introduced questioner-perspective bias score to assess the discrepancy between the accepted answer and the most users-preferred answer, denoted as \(s_q\). 
\begin{equation}
    s_q=\frac{(v-v_a)-v_m}{v_{\sigma } }
\end{equation}
Here, \( v \in \{v_1, \cdots, v_{N_i}\} \) represents the votes for each answer in question \( q \).
\( v_a \) denotes the votes for the answer accepted by the questioner, and \( v_m \) and \( v_{\sigma} \) are the mean and standard deviation of \( \{v_1, \cdots, v_{N_i}\} \), respectively. 

Second, since high-quality text is usually accompanied by a high number of votes \cite{gkotsis2014s}, interaction data from open communities not only reflect the preferences of community users but also serve as a dual filtering mechanism for the quality of the answer content.
Unlike specific answers preferred by the questioner, the number of votes generally reflects the universality and generality of the answer, indicating its applicability and transferability to other contexts. 
This aspect is particularly significant in the field of software engineering, where similar issues may arise in different environments.
The ability to generalize and apply solutions across diverse scenarios is a crucial capability in this field, underscoring the importance of community-driven feedback and the collective wisdom it represents.
Therefore, to comprehensively consider the users' collective preferences and engagement with the answers, we introduce a users-perspective voting score, denoted as \( s_u \), mathematically expressed as follows:
\begin{equation}
s_u = \frac{v - \min(V)}{\max(V) - \min(V)}
\end{equation}
Here, \( v \) belongs to \( V \), and \( V = \{v_1, \cdots, v_{N_i}\} \) is the set of votes for any given answer pool \(\{a_1, a_2, \ldots, a_{N_i}\}\). \(\min(V)\) and \(\max(V)\) represent the minimum and maximum values within \( V \), respectively.
This normalization ensures that the number of votes is adjusted to a common scale, facilitating fair comparison across different answers.

Third, since high semantic accuracy is a fundamental prerequisite for answering questions, we introduce a content score from the perspective of LLMs, denoted as \( s_l \). 
This score aims to leverage LLMs, which excel at handling the nuanced semantic relationships between text and code, to evaluate the quality of text \( c \) in answer \( a \).
The LLMs employed can be general-purpose or specifically tailored for the code domain, typically possessing excellent comprehension and reasoning capabilities. In our study, we selected an easily accessible LLM, denoted as \( \mathcal{CM} \), to measure the semantic logical value of each question-answer pair \( (q_i, a_i) \), where \( a_i \) is an element from the set \(\{a_1^i, a_2^i, \ldots, a_{N_i}^i\}\).
The content score \( s_l \) is calculated by multiplying the probabilities of each token generated by \( \mathcal{CM} \), as detailed in Eq. (\ref{cs}). 
\( \mathcal{I}_1 \) represents a prompt template that integrates the question-answer pair \( (q_i, a_i) \).
\begin{equation}
s_l = \prod_{t \in a_i}\sigma\left(\mathcal{CM}(\mathcal{I}_1[q_i, t])\right)
\label{cs}
\end{equation}
In Eq. (\ref{cs}), the product is taken over all tokens \( t \) in the answer \( a_i \), and \(\sigma\) is the logistic function. 
Ultimately, to comprehensively evaluate the pool of answers and construct a preference sequence, we introduced an overall preference score \( r \), which consolidates various perspectives into a unified measure. 
Subsequently, we ranked the preference set based on the magnitude of \( r \), with the highest-scoring answer \( a \) becoming \( p_1 \).
\begin{equation}
    r = \alpha_1 \cdot s_q + \alpha_2 \cdot s_v + \alpha_3 \cdot s_c
\end{equation}
The weights \( \alpha_1 \), \( \alpha_2 \), and \( \alpha_3 \) reflect the relative importance of each component within the multi-perspective preference set.
\begin{table*}[t]
\caption{ Statistics on the size of the answers pool for each question. }
    \centering
     \renewcommand{\arraystretch}{1.1}
        \tabcolsep=0.4cm
    \begin{tabular}{ccc ccc ccc  }
    \hline
        Count Interval &[0,2) & [2,5) & [5,10)  &  [10,15) & [15,20)&[20,25) &[25,30]& Total  \\ 
         \midrule
        Count & 325,780 & 245,793 & 21,986 & 2,057 &572 & 203 & 222  &596,613\\
        Percentage(\%) & 54.60 & 41.20 & 3.68& 0.35&0.10&0.03&0.04 & 100 \\\hline
       
    \end{tabular}
    \label{data}
\end{table*}
\subsubsection{\textbf{Preference Ranking Alignment}}
Assuming that each answer \( p_i \) in the preference ranking set \(\{p_1, \ldots, p_{N_i}\}\) has been trained to resemble human responses, the RLHF method based on the Bradley-Terry model for pairwise comparisons \cite{ouyang2022training} may be insufficient, such as \(  p_i \succ p_j  \) shown in Figure \ref{method} (b).
To facilitate the comparison of multiple responses, we extended the Bradley-Terry model through Multi-perspective Preference Ranking Alignment (MPRA), inspired by Proximal Policy Optimization (PPO) \cite{schulman2017proximal} and Preference Ranking Optimization \cite{song2023preference}. 
MPRA shifts the focus from a reward model-centric approach to directly adjusting the probability ranking of \( N_i \) answers generated by LLMs to align with the overall preference score \( r \). 
Here, \( N_i \) denotes the size of the answer pool \(\{a_1^i, a_2^i, \ldots, a_{N_i}^i\}\) for question \( q_i \), which varies with the question \( q \).
The comprehensive process of MPRA is illustrated in Figure \ref{method} (c). The initial preference ranking set \(\{p_1, \ldots, p_{N_i}\}\) is given by \( p_1 \succ p_2 \succ \ldots \succ p_{N_i} \), which can be divided into \( p_1 \) and \(\{p_2, \ldots, p_{N_i}\}\). The function parameterized of LLM \(\mathcal{M}_2\) in MPRA is defined as \(\mathcal{M}_2(\mathcal{I}_1,p_k)\).
The extended Bradley-Terry objective is defined as follows:
\begin{equation}
\mathcal{P}(p_{1,2:N_i}|\mathcal{I}_1)= \frac{\exp(\mathcal{M}_2(\mathcal{I}_1, p_1))}{\sum_{k=1}^{N_i} \exp(\mathcal{M}_2(\mathcal{I}_1, p_k))}
\label{round1}
\end{equation}
As the \( i \)-th iteration unfolds, MPRA systematically eliminates the top \( i \) answers with higher \( r \) scores. This process is repeated until all answers are excluded. This iterative refinement continues sequentially until the entire set of potential solutions is exhausted, with the final target probability evolving into:
\begin{align}
\mathcal{P}(p_{1,\ldots,N_i}|\mathcal{I}_1) &= \prod_{i=1}^{N_i-1} \mathcal{P}(p_{i,i+1:N_i}|\mathcal{I}_1) \nonumber \\
&= \prod_{i=1}^{N_i-1} \frac{\exp(\mathcal{M}_2(\mathcal{I}_1, p_i))}{\sum_{k=i}^{N_i} \exp(\mathcal{M}_2(\mathcal{I}_1, p_k))}
\label{roundn}
\end{align}
This optimization objective is intricately aligned with the ultimate aim of human alignment, which is to select the desired response from the expansive response space of LLMs \cite{rafailov2023direct}. Essentially, when \( N_i \rightarrow \infty \), Eq. (\ref{roundn}) is capable of exhaustively exploring all potential responses generated by the LLM; when \( N_i = 2 \), MPRA transforms into the Bradley-Terry model as described in Eq. (\ref{bt}), yet still effectively optimizes the LLM. 
Unlike methods that compel the LLM to approximate a reward model, MPRA directly trains the LLM by optimizing the probability ranking of the preference set through list-wise contrastive learning objectives. Its optimization goal, represented by Eq. (\ref{mpra}), aims to realize Eq. (\ref{roundn}). This process iteratively designates the most favored response with the highest \( r \) score as positive, perfectly aligning with human preferences, while the remaining responses are treated as negative.
MPRA not only needs to generate the most preferred responses but also to enhance text fluency and code structure. Consequently, MPRA incorporates the new SFT stage, as shown in Figure \ref{method} step2 (c), with its loss identical to that of the foundational SFT phase in Eq. (\ref{qpsft}) described in Section \ref{sec:sfp}, denoted as \(\mathcal{L}_{MSFT}\). 
\begin{equation}
\mathcal{L}_{MPRA} = - \sum_{i=1}^{N_i-1} \log \frac{\exp(\mathcal{M}_2(\mathcal{I}_1, p_i))}{\sum_{k=i}^{N_i} \exp(\mathcal{M}_2(\mathcal{I}_1, p_k))}
\label{mpra}
\end{equation}
The crucial difference is that during the foundational SFT phase, the alignment target is the questioner-accepted answer, whereas the LLM’s alignment target now is the answer with the highest overall preference score \( r \). Ultimately, the overall optimization objective can be summarized as follows:
\begin{equation}
\mathcal{L}(p_{1,\ldots,N_i}|\mathcal{I}_1) = \mathcal{L}_{MPRA} + \alpha \mathcal{L}_{MSFT}
\end{equation}
where \(\mathcal{L}_{MSFT}\) is the NLL loss for the top candidate, and \(\alpha\) is a hyperparameter used to balance text quality and human preference.
\subsection{Retrieval-augmented In-context Learning}
As the accepted answer may become outdated with the rapid occurrence of API updates in the field of programming, for instance, in Figure~\ref{intro}, the "urllib" API in answer \(a_2\) is applicable to Python 2 but deprecated in Python 3, we introduced Retrieval-augmented In-context Learning (RIL) to address the issue of outdated answers and to align with the user’s preference for utilizing new API trends. By retrieving analogous questions from the question bank and employing them as few-shot examples, we enhance the efficacy of the generated responses.
We utilize a dense retriever (\(\mathcal{R}_D\)), which excels at handling the transition from natural language to code generation, having been trained to extract documents from a comprehensive pool that includes a vast repository of code libraries, APIs, and functions.
Due to the significant influence of the ordering of few-shot examples on the model’s predictions \cite{zhao2021calibrate}, even though In-context Learning can still perform well when the orders or labels of prompts are exchanged \cite{min2022rethinking}, we select the most similar question-answer \((sq, sa)\) pair from the question-answers bank \( \mathcal{D} \) to serve as the few-shot example in prompt \( \mathcal{I}_2 \). 
The ultimate objective is formulated as follows:
\begin{equation}
    \mathcal{P}(A_t) = \prod_{i=1}^{T} P_{\mathcal{M}_2}(A | \mathcal{I}_2, Q, (sq, sa), A_{<t})
\end{equation}
Here, \( \mathcal{M}_2 \) represents the LLMs following the MPRA stage, and \(Q\) is the question to be resolved
\begin{table*}[t]
  \caption{The zero-shot experimental results on the StaCCQA dataset. Open-source code baselines are above ALMupQA and closed-source baselines are below ALMupQA. The best result in each column is marked in bold. The second-best result in each column is underlined.}
  \label{zero-shot}
  \renewcommand{\arraystretch}{1.1}
  \begin{tabular}{ ccc ccc cc}
    \toprule
   Model& Model size &  \(BLEU_4\)  & \(ROUGE_2\) & \(CHRF\) & \(BERTScore\) & \textit{CodeBERTScore-PR} & \textit{CodeBERTScore-F}\\ 
   \hline
    Godegen-mono &16B&6.72  &9.24  &32.94  &77.53  &54.42 &50.20\\
    GPT-NeoX & 20B &8.40  &11.26  &33.46  &78.06  &53.79  & 49.87\\
    StarCoder& 15B &9.32  &11.92  &30.75  &77.57  &53.36  & 52.21\\
    WizardCoder-python &13B  &12.97 &15.88&37.54 &\underline{79.34} &52.37 & 51.89 \\
    CodeT5+ & - & 3.86 &5.16 &25.58 & 75.96 &53.48 &46.19 \\
   Code Llama2  &7B  &11.86  &16.32& 35.08 & 70.10 & 46.46 & 47.05 \\
    Code Llama2 &13B &13.56  &18.32 &38.68  &78.13 &51.79 &52.91 \\ \hline
   \textbf{ ALMupQA(Ours)} & 7B  &\textbf{22.86} &\textbf{25.48} &\textbf{40.58} &\textbf{84.14} &\textbf{ 65.12} &\textbf{63.53}\\ \hline
    PaLM & -    &13.15 &18.68  &\underline{39.89}  &77.89 &52.81  &51.98\\
    ChatGLM &-  &13.91  &18.71  &38.21  &78.28  &53.29  &\underline{53.77}\\
    GPT-3.5 &-  &\underline{15.29}  &\underline{19.24}  &39.10 &78.90  &52.10 &52.95\\
    Claude2 & - &14.69  &19.12  &38.78  &78.45  &51.58  &52.63\\
    GPT-4 &-  &13.04  &17.74  &35.43  &78.23  &\underline{57.84}  &46.82\\
 
  \bottomrule
\end{tabular}
\end{table*}
\begin{figure}[t]
    \includegraphics[width=\linewidth]{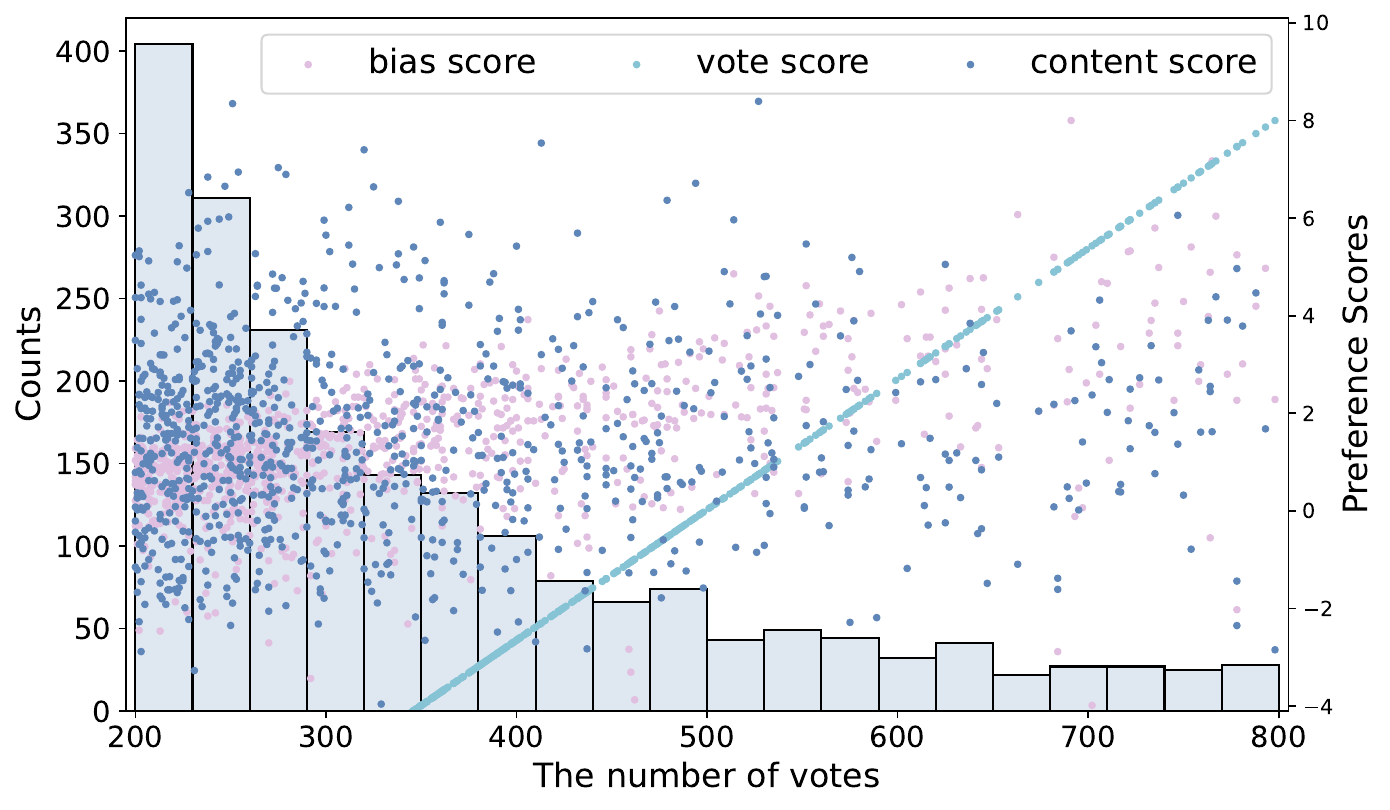}
    \caption{The statistic of the number of votes for a question and the mapping relationship among the bias score \(s_q\), vote scores \(s_u\) and content scores \(s_l\).}
    \label{scores}
    \vspace{-0.3cm}
\end{figure}
\section{EXPERIMENTS}
We examine the performance of our proposed ALMupQA for the CCQA task in this Section. 
In particular, we focus on answering the following research questions:
\noindent\begin{enumerate}[leftmargin=*]
   \item[$\bullet$] \textbf{RQ1}: Can ALMupQA surpass the baseline methods in accuracy metrics within the CCQA task?
    \item [$\bullet$]\textbf{RQ2}: Can each component of ALMupQA make a contribution?
    \item[$\bullet$] \textbf{RQ3}: Can ALMupQA align with human preferences and explore the relationship between preference scores and accuracy metrics?
    \item [$\bullet$]\textbf{RQ4}: Can ALMupQA effectively generate user-centric responses that align with human preferences in the coding community?
\end{enumerate}
\subsection{StaCCQA Dataset Construction}
Due to the lack of high-quality, authentic multi-answer preference datasets in code communities, there is an urgent need to construct a new dataset. To address this gap, we have turned to StackExchange, a platform whose forums are accompanied by rich question-answering metadata information. A publicly available dump of user-contributed content from Stack Overflow, provided by StackExchange under a cc-by-sa 4.0 license, has formed the foundation for the creation of our dataset StaCCQA.

The initial StaCCQA dataset was in XML format, comprising 757,702 \((q_i, a_i)\) pairs, mainly featuring <python> tags, with 600,176 pairs containing code blocks. To obtain the latest answers, we systematically gathered all answers for each question \( q \) on Stack Overflow up to August 2023, resulting in a dataset totaling 596,613 pairs. Detailed statistics are presented in Table~\ref{data}. We then proceeded with the following preprocessing steps and the resulting dataset \(\mathcal{D}\) contains 270,716 \((q_i, \{a_1,\ldots,a_{N_i}\})\) pairs. 
\begin{itemize}[leftmargin=*]
    \item To ensure that submission messages are descriptive, we removed pairs with titles that are shorter than three tokens (including three tokens). This decision follows CCT5 \cite{lin2023cct5}, which stipulates that code comments should contain more than three tokens.
    \item Pairs where the answer did not contain code block content were eliminated to ensure that ALMupQA's reference content includes both text and code, due to the nature of CCQA.
    \item Pairs with an answer pool size smaller than 2 were discarded.
    \item All HTML tags were cleaned and replaced with ``[HTML]'', particularly \texttt{<a href\(\cdots\)>} and \texttt{<img\(\cdots\)>} tags, to ensure the model is not influenced by such exceedingly complex and meaningless content. This decision follows existing research that constructed datasets related to submissions \cite{husain2019codesearchnet,lu2021codexglue}.
\end{itemize}
\begin{table*}
  \caption{The one-shot experimental results on the StaCCQA dataset. The best result in each column is marked in bold. The second-best result in each column is underlined.}
  \label{one-shot}
  \begin{tabular}{ ccc ccc cc}
    \toprule
   Model& Model size &  \(BLEU_4\)  & \(ROUGE_2\) & \(CHRF\) & \(BERTScore\) &  \textit{CodeBERTScore-PR} & \textit{CodeBERTScore-F}\\
    \midrule
    Godegen-mono &16B&8.06  &11.01  &33.32  &78.28  &\underline{54.67} &50.20\\
    GPT-NeoX & 20B &8.95  &11.30  &26.84  &76.68  &52.64  & 51.93\\
    StarCoder& 15B &10.59  &14.40  &33.71  &78.20  &53.43  & 52.96\\
    WizardCoder-python &13B  &13.35 & 15.97 &37.56 &\underline{79.42} &52.70 & 52.11 \\
    CodeT5+ & - & 4.40 &5.60 &25.96 & 75.91 &52.23 &47.52 \\ \hline
       \textbf{ALMupQA (Ours)}&  7B &\textbf{22.86} & \textbf{25.48}& \textbf{40.58}& \textbf{84.14}&\textbf{65.12}&\textbf{63.53}\\
     \midrule
    PaLM & -    &12.77  &18.97  &34.00 &77.90  &52.35  &52.25\\
    ChatGLM &-  &13.47  &17.50  &37.06  & 78.20  & 53.51  &\underline{53.53}\\
    GPT-3.5 &-  &14.50  &18.43  &\underline{39.17} &78.92  &52.64 &52.52\\
    Claude2 & - &14.10  &18.24  &38.25  &78.46  &51.38  &52.36\\
    GPT-4 &-  &\underline{14.73}  &\underline{18.87}  &36.68  &78.78 &52.44  &52.56\\
  \bottomrule
\end{tabular}
\end{table*}
\subsection{Multi-perspective Phenomenon Analysis}
To validate the necessity of multi-perspective preference modeling, we randomly extracted approximately 2,000 entries from the constructed dataset StaCCQA. 
We calculated the bias score, vote score and content score for each answer \(a\). These scores were then mapped onto a two-dimensional coordinate system, as illustrated in Figure~\ref{scores}. The horizontal axis represents the number of votes, the left vertical axis indicates the number of answers, and the right vertical axis denotes the preference score.
Naturally, the vote score \(s_u\) shows a positive correlation with the number of votes. For question audience bias \(s_q\), if the votes \(v\) for a user-favored answer are close to the votes for the answer chosen by the questioner, then \(s_q\) is near the X-axis. In Figure~\ref{scores}, most \(s_q\) values are distant from the X-axis, highlighting a significant divergence between user preferences and the questioner's choices within the coding community.
Analyzing the distribution of content scores \(s_l\): if some answers to question \(q\) are semantically similar, content scores should cluster. The lack of clustering indicates that no single answer comprehensively covers all semantic aspects of the question.

In summary, the observed distinct distributions of content scores, vote scores, and bias scores underscore the existence of diverse preferences from different perspectives. This finding validates the necessity of accurately capturing and presenting user preferences in CCQA, necessitating the adoption of multi-perspective modeling approaches.
\subsection{Experiment Settings}
\subsubsection{Baselines}
The classification of LLMs can be determined by the openness of the technology and whether the code is available for research or commercial use. 
Based on the unique characteristics of CCQA, we selected two types of baseline models. 
The first category consists of general-purpose, closed-source LLMs designed for text generation, including GPT-3.5-turbo, GPT-4 \cite{openai2023gpt4}, PaLM \cite{anil2023palm}, ChatGLM \cite{zeng2022glm}, and Claude2 \cite{anthropic-2023}. 
The second category comprises open-source code LLMs that excel in program synthesis, such as StarCoder \cite{li2023starcoder}, WizardCoder-Python-13B \cite{luo2023wizardcoder}, GPT-NeoX \cite{black2022gpt}, CodeGen-mono-16B \cite{nijkamp2023codegen2}, and Code Llama 2 \cite{roziere2023code}.

\subsubsection{Evaluation Metrics} 
To comprehensively evaluate the experimental results, we employed various evaluation metrics from four perspectives: traditional text generation metrics (BLEU \cite{papineni2002bleu}, Rouge \cite{lin2004rouge}, and CHRF \cite{popovic2015chrf}), model-based metrics (BERTScore \cite{zhang2019bertscore}), code-related metrics (CodeBERTScore \cite{zhou2023codebertscore}), and preference metrics based on GPT-4 evaluations. 
Additionally, considering the similarity between Precision and Recall in CodeBERTScore, we unified these metrics as "CodeBERTScore-PR" (abbreviated as CB-PR). Similarly, the F1 and F-measure in CodeBERTScore were merged into "CodeBERTScore-F" (abbreviated as CB-F).
\begin{table}[t]
  \caption{The ablation study results. We evaluate various stripped-down versions of our model to compare the performance gains brought by different components. The full names of these abbreviations are as follows: SFT (Foundational Supervised Fine-Tuning); MPRA (Multi-perspective Preference Ranking Alignment); \(s_p\) (bias score); \(s_l\) (content scores); \(s_u\) (vote scores); and RIL (Retrieval-augmented In-context Learning). The components in \textbf{bold} have the most significant impact on performance.}
  \label{ablation}
  \renewcommand{\arraystretch}{1.1}
        \tabcolsep=0.1cm
  \begin{tabular}{ lc ccc cc}
    \toprule
   Model&  \(BLEU_4\)  & \(ROUGE_2\) & \(CHRF\) & \(BS\) &  \textit{CB-PR} & \textit{CB-F}\\
   \midrule
   ALMupQA & 22.86  & 25.48  & 40.58  & 84.14  & 65.12 & 63.53\\ 
   \quad w/o SFT   &21.30 &23.62 &37.88 &76.15&59.76&57.73 \\ 
    \quad w/o MPRA  & \textbf{14.62} &\textbf{20.50} &\textbf{39.18} & \textbf{80.41} &\textbf{55.72} &\textbf{53.85} \\
   \quad w/o \(s_q\)   &22.16 &25.18&39.38 &83.34 &64.72 & 62.83 \\
   \quad w/o \(s_u\)  &21.01  &23.23 &40.48  &78.06  &53.79  & 49.87\\
   \quad w/o \(s_l\) &21.56  &24.58  &38.78  &82.54  &64.22 & 62.13\\
    \quad w/o RIL  &21.66  &23.16  &39.18  &81.82  &61.22 &62.53\\
  \bottomrule
\end{tabular}
\end{table}
\subsubsection{Implementation Details}
In this study, we selected Code Llama-Instruct-7B \cite{roziere2023code} as the base model \( \mathcal{M} \), which belongs to a series of large code language models based on Llama 2 \cite{llama2}. 
\begin{figure}[ht]
    \includegraphics[width=0.9\linewidth]{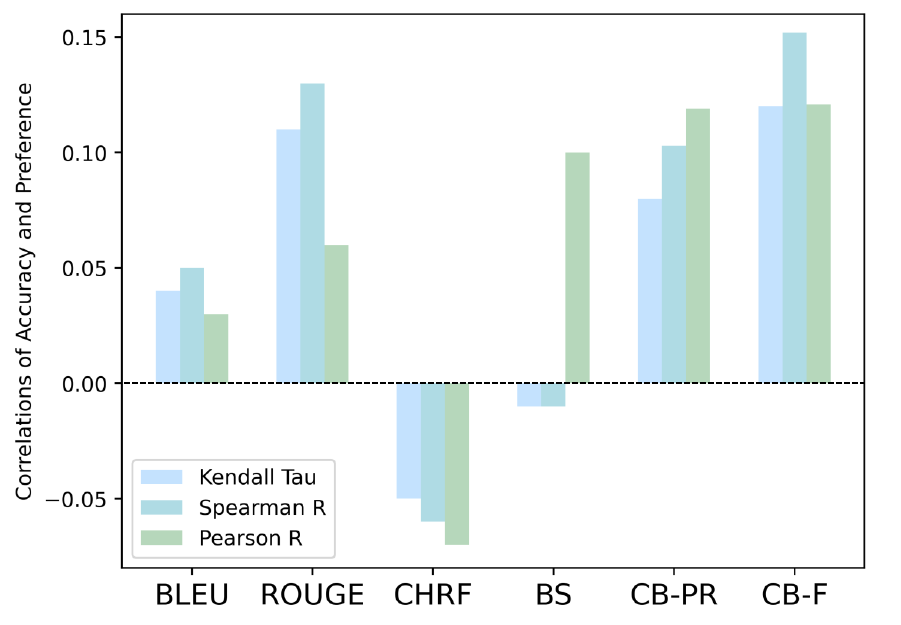}
    \caption{The consistency correlations between accuracy-based metrics (BLEU, ROUGE, CHRF, BERTScore, CB-PR, and CB-F) and preference-based metrics (GPT-4 evaluation scores). A positive correlation indicates that accuracy metrics improve as preference scores increase.}
    \label{corelation}
\end{figure}
Code Llama has demonstrated state-of-the-art performance across various code benchmarks. We utilized the ALMupQA framework for three-stage fine-tuning. During the SFT phase, we specified the following hyperparameters: \(epoch\), \(temperature\), \(top\_p\), \(max\_seq\_len\), and \(max\_batch\_size\), set to 4, 0.2, 0.95, 2048, and 28, respectively. We retained the remaining hyperparameter settings of Llama, which can be found at the following link\footnote{https://github.com/facebookresearch/llama}.
In the MPRA phase, we selected an accessible LLM\footnote{\url{https://huggingface.co/OpenAssistant/oasst-rm-2-pythia-6.9b-epoch-1}} as \( \mathcal{CM} \). The hyperparameters were set as follows: \(learning\_rate\), \(gradient\_accumulation\_steps\), \(epochs\), \(top\_p\), \(max\_gen\_len\), \(temperature\) and \(max\_batch\_size\), set to 1e-4, 9, 4, 1.0, 0.95, 512, and 4, respectively. 
Given the excellent performance of this retrieval-generation approach in understanding diverse texts and code, during the Retrieval-augmented In-context Learning (RIL) phase, we chose the DocPrompting method based on SimCSE \cite{zhou2022docprompting} as our retriever \(\mathcal{R}_D\). 
This retriever \(\mathcal{R}_D\) includes 35,763 functions from all Python libraries on DevDocs\footnote{\url{https://devdocs.io}}, encompassing the Python standard library and widely-used packages such as NumPy and Pandas, and was pre-trained on the re-split CoNaLa \cite{yin2018learning} benchmark.
\begin{figure*}[t]
    \includegraphics[width=0.98\linewidth]{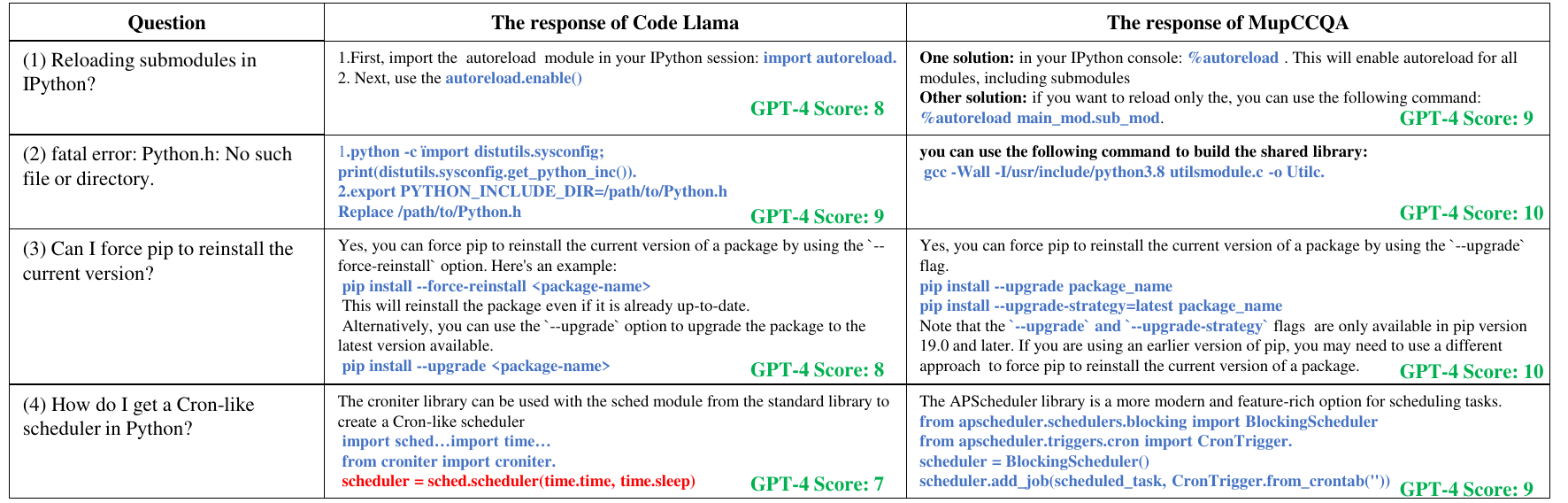}
    \caption{A case study on the performance of Code Llama and ALMupQA in Code Community Question Answering. The black text signifies natural language. The blue text represents programming block. The red text indicates outdated content. The green indicates the preference score based on GPT-4 evaluation. The higher the score, the greater the alignment with user preferences. The scores ranges from [1,10].}
    \label{case}
\end{figure*}
\subsection{RQ1: Main Results}
In this experiment, we used text and code metrics to evaluate the performance of general baselines and code baselines on the StaCCQA dataset, and compared the results. The Table ~\ref{zero-shot} below summarizes the experimental outcomes, presenting the following observations:

First, our ALMupQA significantly outperformed other baseline models across all metrics. Specifically, ALMupQA exceeded the next best results in every metric, surpassing general LLMs in text generation and code LLMs in code generation. For instance, compared to the second-best GPT-3.5, it scored 5.2\% higher on BERTScore, nearly 7\% higher on CB-PR, and about 10\% higher on CB-F than ChatGLM. This indicates that ALMupQA is a robust and versatile model, excelling in the quality, grammar, and semantics of generated answers. Notably, general large language models might have superior code generation capabilities in the CCQA task compared to code LLMs.
Second, ALMupQA showed significant improvements compared to the baseline models, even doubling the BLEU score of the benchmark model (22.86\% vs. 11.86\%). Other n-gram-based metrics (ROUGE and CHRF) and semantic-grammar-based metrics also saw substantial enhancements, indicating that ALMupQA is an effective framework for applying large models to CCQA problems.

Due to the presence of similar question-answer pairs as few-shot examples in the Retrieval-Enhanced In-Context Learning (RIL) of ALMupQA, we also applied RIL to the remaining baselines in a new experiment, and the results are presented in Table~\ref{one-shot}.
This aims to ensure fairness in comparing other zero-shot baselines with ALMupQA. In Table~\ref{one-shot}, each baseline shows improvement across various metrics compared to the zero-shot results in Table~\ref{zero-shot}, with GPT-4 exhibiting significant enhancement in long-text performance, becoming the second-best baseline. 
However, they still cannot match our ALMupQA. 
Specifically, the BLEU score of ALMupQA remains far higher than the second-best GPT-4; the BERTScore of ALMupQA also surpasses the  WizardCoder-python by nearly 4.7\%. 
In terms of CB-PR and CB-F, ALMupQA exceeds the second-best baseline by nearly 10\%.

\subsection{RQ2: Ablation Study}
To validate the enhancement in performance brought by MuCCQA's preference scores across three stages and three different perspectives scores, we conducted ablation experiments. The results are shown in Table~\ref{ablation}.
Upon removing SFT, all metrics experienced a decline, with BERTScore showing the most significant drop (from 84.14\% to 76.15\%), underscoring the importance of this stage for understanding the semantics of programming domain knowledge.
Eliminating MPRA resulted in notable decreases in model performance on metrics focused on complex phrase matching and code semantics, specifically CodeBERTScore, which dropped by 8.8\% and 9.4\%, respectively. 
This suggests that an unadapted LLM fails to account for the diversity of preferences within the coding community.
If RIL is excluded, MuCCQA's performance on semantic-focused metrics (BERTScore and CodeBERTScore) significantly declines, highlighting the critical role of similar examples in understanding problem semantics.
Furthermore, all three preference scores in MuCCQA, including bias scores \(s_q\), vote scores \(s_v\), and content scores \(s_c\), contribute to its performance. In summary, each component of our method plays a unique role, collectively enhancing overall performance.

\subsection{RQ3: GPT-4 Evaluation}
Given that GPT-4 \cite{openai2023gpt4} has demonstrated significant ability in effectively evaluating question-and-answer pairs and aligning with human preferences \cite{wang2023large, zheng2023judging}, we utilize it to assess the preferences for responses generated by the open-source Code Llama, the closed-source GPT-3.5, and our proposed ALMupQA.
To evaluate whether the responses align with human preferences, we designed evaluation criteria encompassing four dimensions: the usefulness, relevance, accuracy, and level of detail of each answer. Each solution is rated on a scale from 1 to 10, with comprehensive explanations required for each score. We need to provide GPT-4 with the question title and specific description, the standard answer, and the responses generated by the LLMs to be evaluated.
The GPT-4 evaluation results indicate that ALMupQA is capable of generating responses that are highly aligned with human preferences, with an average score surpassing that of GPT-3.5 (7.54 to 7.51) and Code Llama (7.54 to 7.43).

To explore the consistency between accuracy-based metrics (including BLEU, ROUGE, CHRF, BERTScore, CB-P, and CB-F) and preference-based metrics (GPT-4 evaluated preference scores), we employed three key statistical correlation coefficients: Kendall's Tau \(\tau\) \cite{kendall1938new}, Spearman's R \(\gamma\) \cite{pranklin1974introduction}, and Pearson's R \(\rho\) \cite{bravais1844analyse}, as depicted in Figure~\ref{corelation}.
The Figure~\ref{corelation} primarily illustrates three points: 
First, the three correlation measures,\(\tau\), \(\gamma\), and \(\rho\), maintain a high degree of sign consistency. 
Second, text-based metrics (BLEU and ROUGE), semantic-based metrics (BERTScore), and code-based metrics (CB-P and CB-F) all exhibit a positive correlation with preference-based metrics, whereas CHRF shows a negative correlation in both \(\tau\) and \(\gamma\).
Lastly, the correlation between code-based metrics and preference is the most pronounced, which aptly reflects the characteristics of our code community question-answering tasks. Overall, accuracy and preference are not contradictory, providing a valuable reference for evaluating CCQA tasks.
\subsection{RQ4: Case Study}
To validate the excellence of our ALMupQA, we selected four random questions for comparison, as shown in Figure \ref{case}. To analyze whether ALMupQA effectively generates user-centric responses, we take randomly the third question, "How to force pip to reinstall the current version?" as an example.
ALMupQA scored 10, while Code Llama scored 8. Although both responses covered the core points and clearly explained how to use the "\mydash force\textendash reinstall" flag, ALMupQA excelled in the following aspects:
First, in detail: ALMupQA provided a more thorough explanation, covering not only the "\mydash upgrade" flag but also the "\mydash upgrade\textendash strategy" flag. This additional information helped users understand and manage package upgrades better.
Second, in accuracy and relevance: ALMupQA accurately explained the usage of the "\mydash upgrade" and "\mydash upgrade\textendash strategy" flags, making the response more informative and helpful for managing package versions and upgrades.
Third, in user-friendliness: ALMupQA's response was well-structured and user-friendly, with clear instructions and examples that made it easier for users to follow and apply the information.

The fourth question in the Figure~\ref{case} aims to highlight the presence of outdated APIs in some responses generated by LLMs. 
Specifically, Code Llama employed the "sched" module, which is part of the Python standard library, but is no longer as commonly used. 
In contrast, the response of ALMupQA utilized a more contemporary library "APScheduler", a popular and feature-rich option for scheduling tasks.
\section{CONCLUSION}
In this paper, aming to explore the application of LLMs with RLHF for human preference alignment in programming-domain Code Community Question Answering (CCQA), we propose ALMupQA.
First, we introduce a multi-perspective preference ranking alignment framework to accommodate diverse user preferences.
Second, to address users' inclination towards using new APIs, we implement a retrieval-augmented in-context learning (RIL) module to mitigate the issue of outdated information.
We conducted extensive experiments to validate the accuracy of our ALMupQA-responsed answers on our crafted dataset StaCCQA, demonstrating an improvement of nearly 11\% in the BLEU compared to the foundation model, with increases of 20\% and 17.5\% in BERTScore and CodeBERTScore, respectively.
Additionally, GPT-4 evaluations confirmed the increase in accuracy-based metrics, with preference scores also showing improvement, indicating the effectiveness of our approach in aligning preferences in the CCQA task.
Overall, we emphasize a novel perspective that considers the diversity of users when aligning with human preferences.
\bibliographystyle{ACM-Reference-Format}
\bibliography{main.bib}

\end{document}